\documentclass[letterpaper, 10 pt, conference]{ieeeconf}  

\IEEEoverridecommandlockouts                              

\usepackage{graphicx} 
\usepackage{epsfig} 
\usepackage{amsmath} 
\usepackage{amssymb}  
\usepackage[T1]{fontenc}
\usepackage{amsfonts} 
\usepackage{import}
\usepackage{xcolor}
\usepackage{float}
\usepackage{cite}
\usepackage[caption=false,font=footnotesize]{subfig}

\newtheorem{theorem}{Theorem}

\newtheorem{lemma}{Lemma}
\newtheorem{remark}{Remark}
\newtheorem{problem}{Problem}
\newtheorem{example}{Example}

\DeclareMathOperator*{\argmin}{argmin}

\title{\LARGE \bf
Least Restrictive Halfplane Control Barrier Functions \\
for an Autonomous Hydrofoil Boat with Moving Obstacles
}

\author{Mattias Trende and Petter Ögren
\thanks{The authors are with the Robotics, Perception and Learning Lab., School of Electrical Engineering and Computer Science, Royal Institute of Technology (KTH), SE-100 44 Stockholm, Sweden, {\tt\small \{mtrende, petter\}@kth.se}}%
}

\begin{document}

\maketitle
\thispagestyle{empty}
\pagestyle{empty}

\begin{abstract}

Finding valid Control Barrier Functions (CBFs) is difficult for non-linear systems subject to actuation constraints and moving in complex environments.
One way to enable analytically derived CBFs is to represent obstacles as supporting halfplanes, allowing the agent to move safely on one side.
In this paper, we propose such a CBF for an agent with Dubins car dynamics.
A potential drawback of halfplane CBFs is that they can be highly conservative, overestimating the obstacles and thereby restricting the agent's motion.
To address this problem, we make the CBF halfplanes less restrictive as follows.
First, the orientation of the supporting halfplane for a single obstacle is optimised to find a safe control input as close as possible to the desired one.
Second, we use multiple halfplanes to better approximate complex geometries, and derive rules for handling halfplane intersections.
Comparative simulations show that our method reduces conservatism compared to existing CBF techniques, enabling tighter obstacle clearance and faster goal arrival, with delays reduced by more than 33\%.
Finally, experiments on an autonomous hydrofoil boat demonstrate successful real-world implementation.


\end{abstract}

\section{Introduction}


Despite advances in maritime perception and situational awareness, including cameras, radars, and AIS transponders, serious accidents still occur \cite{valberg_bilderna_2026}. 
This holds whether a human is in the loop, such as the captain of a small boat, or control is automated, such as an autopilot on a merchant ship.
The problem is not only the lack of information itself, but also acting on it to ensure safety.
In this paper, we design safe control for an autonomous hydrofoil boat, but much of the work will translate to other non-holonomic vehicles such as normal boats, cars, and fixed-wing aircraft.

A popular approach for safe control is based on Control Barrier Functions (CBFs) \cite{ames_control_2019}. The main benefits of using CBFs are that they ensure provable safety while being computationally lightweight. However, CBFs also come with challenges, as it can be difficult to find a valid CBF for high-dimensional systems with actuation constraints \cite{chen_backup_2021}. A common solution is to use Higher Order CBFs (HO-CBFs) \cite{xiao_high-order_2022, haraldsen_hybrid_2025}, but purely distance-based designs require careful tuning to ensure both safety and actuation constraints \cite{kim_is_2026}.

\begin{figure}
    \centering
    \def\svgwidth{0.49\textwidth}\footnotesize
\begingroup%
  \makeatletter%
  \providecommand\color[2][]{%
    \errmessage{(Inkscape) Color is used for the text in Inkscape, but the package 'color.sty' is not loaded}%
    \renewcommand\color[2][]{}%
  }%
  \providecommand\transparent[1]{%
    \errmessage{(Inkscape) Transparency is used (non-zero) for the text in Inkscape, but the package 'transparent.sty' is not loaded}%
    \renewcommand\transparent[1]{}%
  }%
  \providecommand\rotatebox[2]{#2}%
  \newcommand*\fsize{\dimexpr\f@size pt\relax}%
  \newcommand*\lineheight[1]{\fontsize{\fsize}{#1\fsize}\selectfont}%
  \ifx\svgwidth\undefined%
    \setlength{\unitlength}{566.92913386bp}%
    \ifx\svgscale\undefined%
      \relax%
    \else%
      \setlength{\unitlength}{\unitlength * \real{\svgscale}}%
    \fi%
  \else%
    \setlength{\unitlength}{\svgwidth}%
  \fi%
  \global\let\svgwidth\undefined%
  \global\let\svgscale\undefined%
  \makeatother%
  \begin{picture}(1,0.65)%
    \lineheight{1}%
    \setlength\tabcolsep{0pt}%
    \put(0,0){\includegraphics[width=\unitlength,page=1]{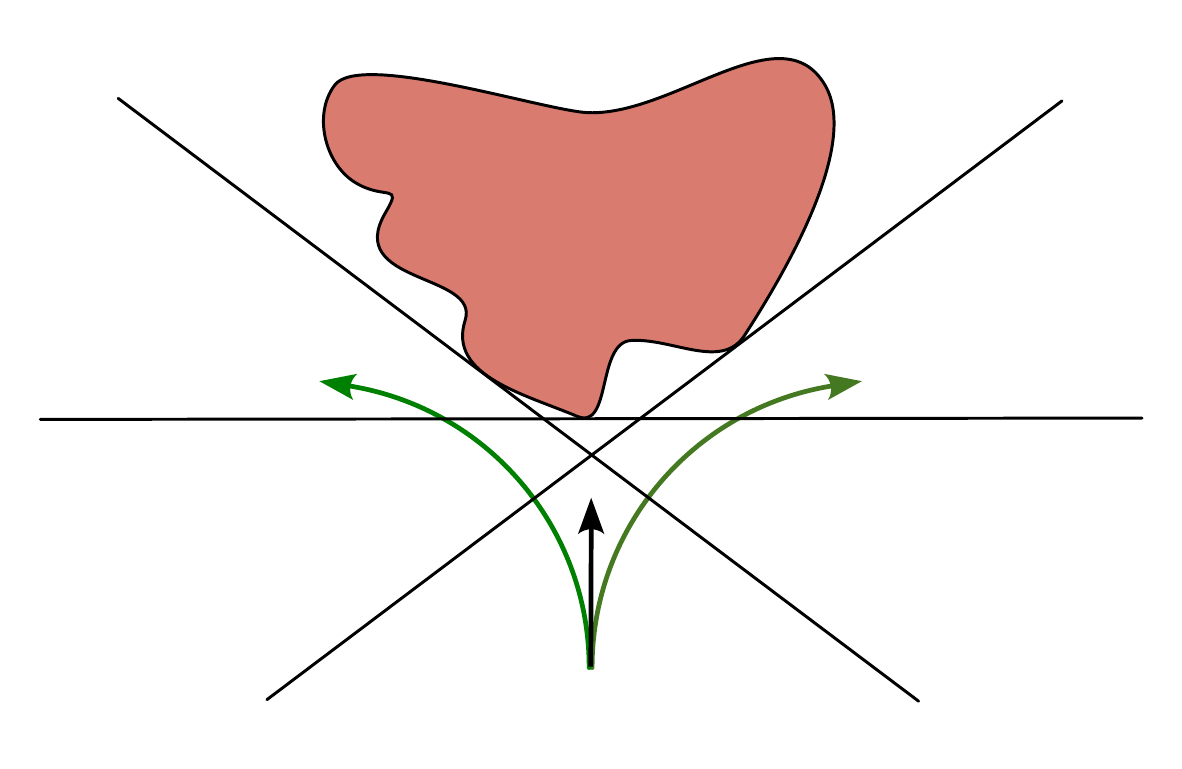}}%
    \put(0.52559255,0.08338839){\color[rgb]{0,0,0}\makebox(0,0)[lt]{\lineheight{1.25}\smash{\begin{tabular}[t]{l}$A$\end{tabular}}}}%
    \put(0.45330115,0.39858929){\color[rgb]{0,0,0}\makebox(0,0)[lt]{\lineheight{1.25}\smash{\begin{tabular}[t]{l}$O$\end{tabular}}}}%
    \put(0.80440994,0.06293904){\color[rgb]{0,0,0}\makebox(0,0)[lt]{\lineheight{1.25}\smash{\begin{tabular}[t]{l}$h_3$\end{tabular}}}}%
    \put(0.80440994,0.26365182){\color[rgb]{0,0,0}\makebox(0,0)[lt]{\lineheight{1.25}\smash{\begin{tabular}[t]{l}$h_2$\end{tabular}}}}%
    \put(0.80440994,0.46436455){\color[rgb]{0,0,0}\makebox(0,0)[lt]{\lineheight{1.25}\smash{\begin{tabular}[t]{l}$h_1$\end{tabular}}}}%
    \put(0,0){\includegraphics[width=\unitlength,page=2]{Intro.pdf}}%
    \put(0.81655746,0.59344118){\color[rgb]{0,0,0}\makebox(0,0)[lt]{\lineheight{1.25}\smash{\begin{tabular}[t]{l}$G_2$\end{tabular}}}}%
    \put(0.13337379,0.58550369){\color[rgb]{0,0,0}\makebox(0,0)[lt]{\lineheight{1.25}\smash{\begin{tabular}[t]{l}$G_1$\end{tabular}}}}%
  \end{picture}%
\endgroup%

    \caption{
    A boat approaching a stationary obstacle (moving obstacles will be considered below).
    Three different supporting halfplane CBFs are shown as $h_1, h_2$, and $h_3$.
    The green curves correspond to the boat's turning radius.
    For the CBF $h_2$, which is orthogonal to the minimum distance between the boat and the obstacle, the boat is not in the safe set, as neither turning direction can avoid the halfplane.
    By either choosing $h_1$ or $h_3$, we can ensure safety.
    Depending on the boat's destination ($G_1$ or $G_2$), the least restrictive halfplane CBF will either be $h_1$ or $h_3$.
    }
    \label{fig:front_page}
\end{figure}

Previous studies often construct CBFs using the signed distance function (SDF) for non-holonomic systems like Dubins cars, unicycles, and bicycles \cite{marley_maneuvering_2021, shaw_cortez_robust_2022, haraldsen_hybrid_2025}. 
To handle the second-order dynamics, one can apply variants of Higher-Order CBFs (HO-CBFs) as in \cite{marley_maneuvering_2021, shaw_cortez_robust_2022, haraldsen_hybrid_2025}, or include a term containing the relative velocity between the agent and the obstacle \cite{thyri_domain-based_2022, basso_safety-critical_2020}.

To investigate the relationship between SDF CBFs and a class of CBFs based on halfplanes,
we note that the distance from a vehicle to the closest point on an obstacle is the same as the distance to a supporting halfplane through that point, perpendicular to the line between the point and the vehicle.
Thus, the SDF CBFs are equivalent to
Minimum Distance Halfplane CBFs (MDH-CBFs), and to underline this connection, we adopt the MDH-CBF notation for the remainder of this paper.

Moving away from the MDH-CBFs, \cite{huang_dynamic_2025} adds a velocity component to the CBF that is derived from a velocity obstacle. None of the papers \cite{marley_maneuvering_2021, shaw_cortez_robust_2022, haraldsen_hybrid_2025, thyri_domain-based_2022, basso_safety-critical_2020, huang_dynamic_2025} explicitly consider actuation constraints when designing the CBF function, instead compliance is achieved through parameter tuning. One that does include constraints is \cite{squires_constructive_2018}, which looks forward in time to determine the value of the CBF. This enables the CBF to handle actuation constraints at the price of higher computational costs.

\begin{figure}[!t]
    \centering
    \includegraphics[width=0.99\columnwidth]{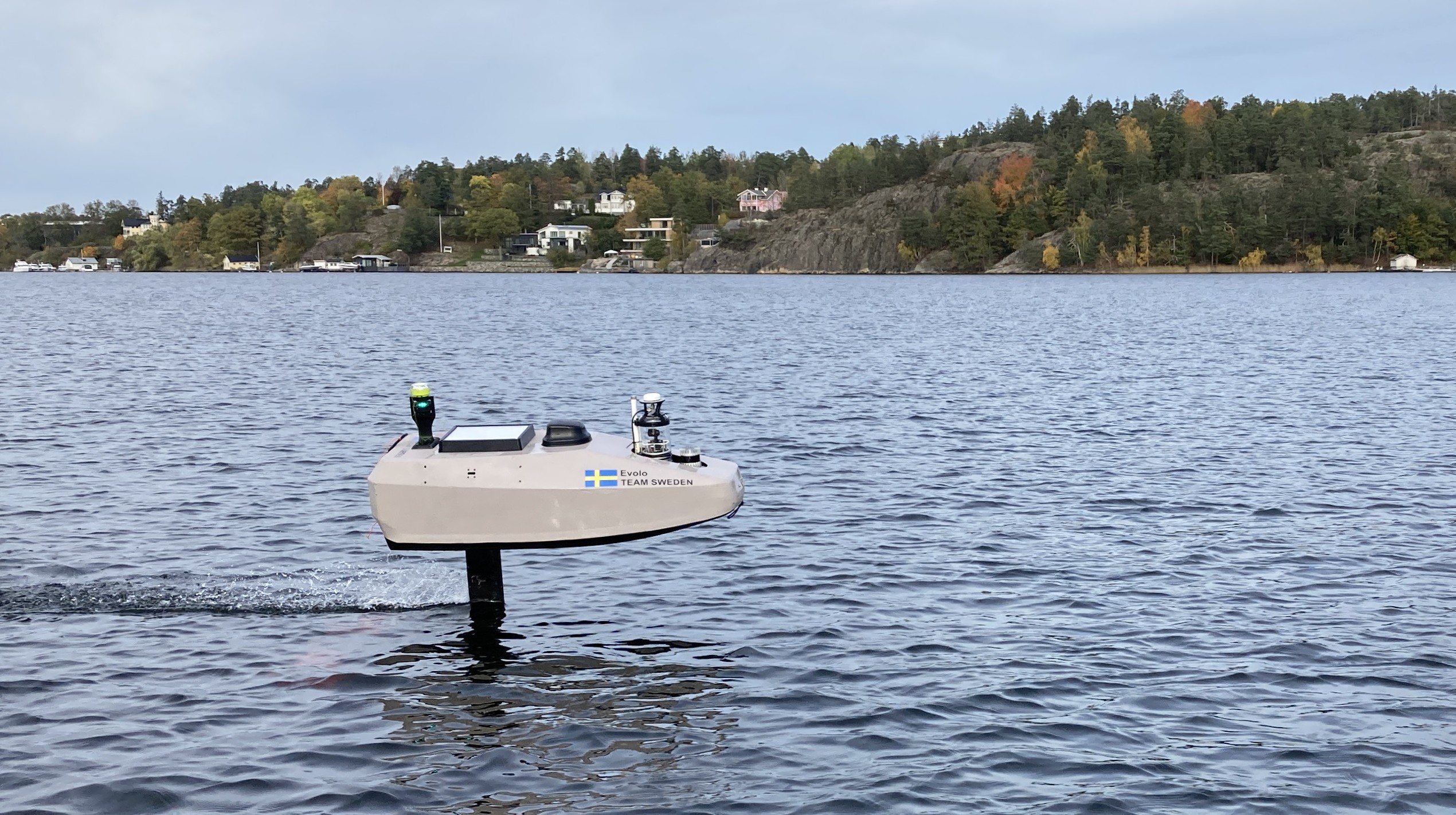}
    \caption{The autonomous hydrofoiler, Evolo.}
    \label{fig: evolo pic}
\end{figure}

In this paper, we will study CBFs for the Dubins car model \cite{dubins_curves_1957}. This is a fairly accurate model for our
electric hydrofoil boat Evolo, see Figure \ref{fig: evolo pic}, which has a limited turn radius and a narrow speed regime with high energy efficiency, motivating the fixed speed assumption. The Dubins car model ignores the tilting dynamics and the hydrofoil control surfaces, but describes the behaviour well at larger time scales.

Building upon Dubins car model, we propose an analytical CBF formulation similar in structure to the analytical solutions for the double integrator found in \cite{borrmann_control_2015, chen_backup_2021}. Intuitively, we replace the braking distance derived for the double integrator with the turning behaviour of Dubins car, ensuring that it always has enough space to perform a turn without hitting the moving obstacle halfplane, up until the point when the closing distance to the moving halfplane is zero. 

As noted above, halfplane obstacle approximations tend to be conservative. To reduce this effect,
building upon our previous work in \cite{trende_least_2026}, we optimise over the orientation of the supporting halfplane.
Ideally, we find an orientation that makes our desired control safe, and if that is not possible, we minimize the gap between the safe and the desired control. We call this the \emph{least restrictive} halfplane CBF (LRH-CBF), as illustrated in Figure \ref{fig:front_page}. For non-convex obstacle configurations, it can be beneficial to combine several halfplanes, as discussed in Section \ref{subsection: Multiple halfplanes}.

The contribution of this paper consists of two main parts. First, the closed-form CBF for Dubins car and the handling of multiple moving obstacles and non-convex obstacle geometries. Second, the hardware experiment and the batch simulations show how the proposed approach leads to less conservative trajectories, making smaller detours that reach the target state in shorter time.

\section{Background}
\label{section: Background}
In this section, we will describe the Dubins car model, CBFs in general, and least restrictive CBFs in particular.
\subsection{Dubins Car}

Dubins car models a vehicle that travels at a constant speed $v$, being controlled by the bounded angular velocity $\omega$ \cite{dubins_curves_1957}. The system state is given by $\mathbf{x} = [x, y, \phi]^T$, where $x$ and $y$ denote the position, and $\phi$ the heading. The continuous-time dynamics in control-affine form are given by
\begin{equation} \label{eq: unicycle dyn}
    \dot{\mathbf{x}} = 
        \begin{bmatrix}
            v \ \cos(\phi) \\
            v \ \sin(\phi) \\
            0 
        \end{bmatrix}
        + 
        \begin{bmatrix}
            0\\
            0\\
            1
        \end{bmatrix} \omega.
\end{equation}
The input is limited to $|\omega| \leq \omega_m$, where $\omega_m$ is the maximum angular velocity. The minimum turning radius then becomes
\begin{equation} \label{eq: turn radius}
    r_m = \frac{v}{\omega_m}.
\end{equation}

\subsection{Control Barrier Functions}

Following the CBF definition in \cite{ames_control_2019}, consider a nonlinear affine control system in the form 
$\dot{\mathbf{x}} = \mathbf{f}(\mathbf{x}) + \mathbf{g}(\mathbf{x})\mathbf{u},$
where $\mathbf{f}$ and $\mathbf{g}$ are locally Lipschitz, $\mathbf{x} \in \mathbb{R}^n$ is the system state, and $\mathbf{u} \in \mathbb{R}^m$ is the system input. 
Then introduce the safe set $\mathcal{C} \subset \mathbb{R}^n$ given by $\mathcal{C} = \{\mathbf{x} \in \mathbb{R}^n \ | \ h(\mathbf{x} ) \geq 0\}$ for some continuously differentiable function $h:\mathbb{R}^n \to \mathbb{R}$.
For safety, $\mathcal{C} \cap O = \emptyset$ must hold, where $O$ is the obstacle set.
If $h(\mathbf{x)}$ also satisfies the inequality 
\begin{equation} \label{eq: h_lim}\sup_{\mathbf{u} \in \mathcal{U}}
    \dot{h}(\mathbf{x},\mathbf{u}) \geq -\alpha(h(\mathbf{x})),
\end{equation}
for all $\mathbf{x} \in \mathcal{C}$, then $h$ is a CBF. Here $\alpha : \mathbb{R} \to \mathbb{R}$ is an extended class $\mathcal{K}_\infty$ function, i.e., $\alpha$ is strictly increasing and satisfies $\alpha(0) = 0$ and $\lim_{x\to\pm\infty}\alpha(x) = \pm \infty$.

Given a desired control $\mathbf{u}_{\text{des}}$, the search for a safe control $\mathbf{u}$ close to the desired one can be formalised in the following optimisation problem \cite{ames_control_2019}:

\begin{problem}[CBF-QP]
\label{prob_CBF_QP}
\begin{align}
    \mathbf{u}(\mathbf{x}) = \argmin_\mathbf{u} & \ (\mathbf{u}-\mathbf{u}_{\text{des}})^T \mathbf{Q} (\mathbf{u}-\mathbf{u}_{\text{des}}) \\
    \text{s.t. } 
    & \ 
    \dot{h}(\mathbf{x},\mathbf{u}) \geq -\alpha(h(\mathbf{x})) \label{eq: cont cons CBF}\\
    & \ \mathbf{u} \in \mathcal{U},
\end{align}
where $Q$ is a positive definite matrix.
\end{problem}

\subsection{Least Restrictive Halfplane CBFs}

The orientation of a halfplane CBF can greatly influence how restrictive it is in terms of the control deemed safe, and thus how conservative the trajectories are. In \cite{trende_least_2026}, double integrator systems are studied, where by optimising the halfplane orientation, the LRH-CBF is found in relation to the desired control input. Parametrising the halfplane normal vector by $\theta$, i.e., $\hat{n}(\theta) = [\cos \theta, \sin \theta]$ and including this parameter in the optimization of Problem \ref{prob_CBF_QP} results in the LRH-CBF Optimisation Problem:

\begin{problem} [LRH-CBF-OP]
\label{prob_LRH_CBF_OP}
    \begin{align}
        (\mathbf{u}(\mathbf{x}), \theta) = \argmin_{\mathbf{u},\theta} & \ (\mathbf{u}-\mathbf{u}_{\text{des}})^T \mathbf{Q} (\mathbf{u}-\mathbf{u}_{\text{des}}) \\
        \text{s.t. } 
        & \ 
        \dot{h}(\mathbf{x},\theta,\mathbf{u}) \geq -\alpha(h(\mathbf{x},\theta)) \label{eq_CBF_constr}\\
        & \
        {h}(\mathbf{x},\theta) \geq 0  \label{eq_feasible}\\
        & \        
        \mathbf{u} \in \mathcal{U}, \ \theta \in \Theta.  \label{eq: CBF ineq} 
    \end{align}
\end{problem}

Thus, by optimising over both $\theta$ and $\mathbf{u}$, the least restricted safe control is found.

\begin{remark}
    By generalising Problem \ref{prob_CBF_QP} to Problem \ref{prob_LRH_CBF_OP}, the optimisation is no longer a QP, as the constraints become non-linear and non-convex in $\theta$. However, note that for any fixed $\theta$, it once again becomes a QP, for which very efficient solvers exist. Hence, we
    can choose $\Theta$ to be a finite set (of about 8-20 elements) and then solve the QP for each element of this set to find the optimal solution $(\mathbf{u},\theta)$,
    \cite{trende_least_2026}.
\end{remark}

\section{Proposed Approach}
\label{section: Method}

First, a halfplane CBF for Dubins car is presented, under the assumption that the halfplane moves with constant velocity. It is proven to be both safe and minimally conservative. Then, the special case when the halfplane is stationary is stated, yielding a simplified CBF. Finally, a method for handling multiple CBF halfplanes is presented, which is useful for non-convex obstacles and multi-obstacle settings.

\subsection{CBF for Dubins Car with moving obstacles}

\begin{figure}[t!]
    \centering
    \def\svgwidth{0.49\textwidth}\footnotesize
    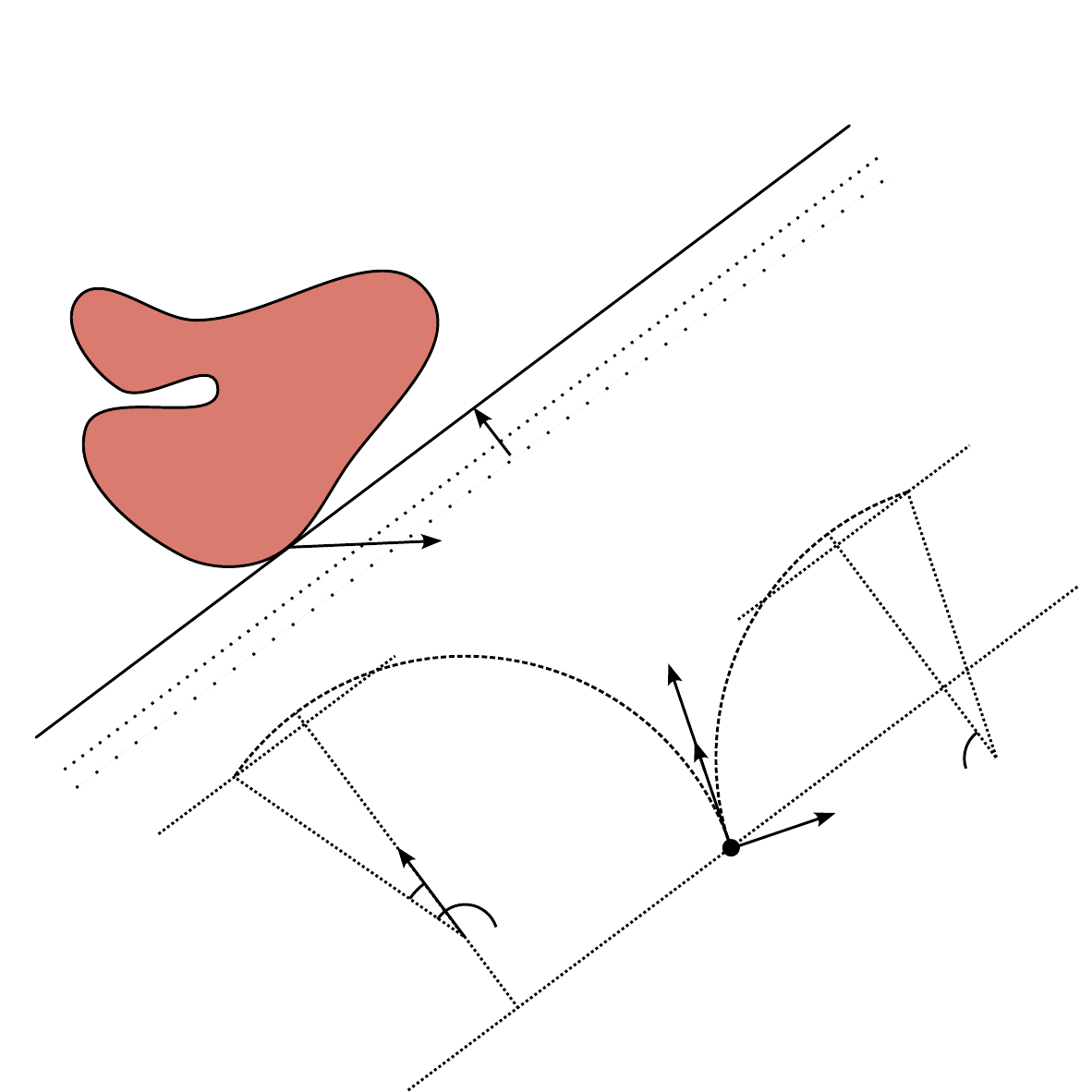
    \caption{The geometry of a Dubins car agent, $A$, approaching a moving halfplane, $O$. Two circular arc turns are shown through angles $\beta_{1}$ (left turn) and $\beta_{-1}$ (right turn). The two terms of $b$ in \eqref{eq: b for moving h} are divided into $b_a$ and $b_b$ in the figure.}
    \label{fig: move H geometry}
\end{figure}

We will now present the CBF for Dubins car. The geometry is shown in Figure~\ref{fig: move H geometry}, including many of the expressions used in the derivation.

\begin{theorem}
\label{theorem: move}
    Consider a moving halfplane $O$, defined by a point in the plane $\mathbf{p}_O$ with constant velocity $\mathbf{v}_O$ and the normal vector $\hat{\mathbf{n}}(\theta)=[\cos \theta, \sin \theta]^T$, and an agent $A$ with Dubins car dynamics given by \eqref{eq: unicycle dyn}.
    For each turning direction $\mu \in \{-1, 1\}$ (where $\mu = 1$ is a left turn, and $\mu = -1$ is a right turn), a CBF is given by 
    \begin{equation}\label{eq: move CBF}
        h_\mu(\mathbf{x}, \theta) = \hat{\mathbf{n}}(\theta)^T(\mathbf{p}_A - \mathbf{p}_O) - \delta - b,
    \end{equation}
    where
    \begin{equation} \label{eq: b for moving h}
        b = r_m (\cos(\gamma) - \cos(\beta_{\mu})) + \frac{v_H r_m (\beta_{\mu} + \gamma)}{v_A}.
    \end{equation}
    Furthermore,
    \begin{align}
        v_H &= \hat{\mathbf{n}}(\theta)^T \mathbf{v}_O,\\
        \gamma &= \text{arcsin}\left(\frac{v_H}{v_A}\right),\\
         \beta_{\mu} &= (\mu (\theta - \phi) - \pi/2), 
    \end{align}
    where $\beta_\mu \in [-\gamma, 2 \pi - \gamma)$.
\end{theorem}

Before stating the proof, we give a geometric interpretation of the expressions above, looking at Figure~\ref{fig: move H geometry}.
The first term $\hat{\mathbf{n}}(\theta)^T(\mathbf{p}_A - \mathbf{p}_O)$ is the distance to the halfplane, $\delta$ is the agent radius, and $b$ is the margin required for the evasive maneuver, ending in a safe heading, in the sense of keeping a constant distance to the plane.
The first term in $b$, $b_a = r_m (\cos(\gamma) - \cos(\beta_{\mu}))$, is the minimal distance in the normal direction of the halfplane it takes to perform a minimum-radius avoidance turn to get to the safe heading. 
Here $\beta_\mu$ is the turning angle needed to move parallel to the halfplane, if turning in direction $\mu$, while $\gamma$ is the heading angle, measured from the halfplane tangent, needed to keep a constant distance to the plane.
The second term in $b$, $b_b = v_H r_m (\beta_{\mu} + \gamma)/v_A$, is the distance the halfplane will move, in the direction of $\hat{n}(\theta)$, during the execution of the turn calculated in $b_a$.
Here, $v_H$ is the speed of the halfplane in the direction $\hat{n}(\theta)$. Note that we assume $v_H \leq v_A$, otherwise, there is no escape from the moving halfplane.

\begin{proof}
     To show that the CBF $h_\mu$ in \eqref{eq: move CBF} is indeed a valid CBF, it must be shown that the condition \eqref{eq: h_lim} holds for all $\mathbf{x} \in \mathcal{C}_\mu$, where $\mathcal{C}_\mu:= \{\mathbf{x}: h_\mu(\mathbf{x}, \theta) \geq 0\}$. Start by noting that $-\alpha(h_\mu(\mathbf{x}, \theta)) \leq 0$ for all $\mathbf{x} \in \mathcal{C}_\mu$, since $h_\mu(\mathbf{x}, \theta) \geq 0$ and $\alpha \in \mathcal{K}_\infty$.

     Thus, if we show that 
$\sup_{\mathbf{u} \in \mathcal{U}} \dot{h}(\mathbf{x},\mathbf{u}) =0 $, then \eqref{eq: h_lim}  holds.
$\dot{h}(\mathbf{x},\mathbf{u})= \nabla h \dot x+ \frac{\partial h}{\partial t} $, so we start with $\nabla h$.
\begin{align}
    \nabla h =& \begin{bmatrix}
        \frac{\partial h}{\partial x} \\
        \frac{\partial h}{\partial y} \\
        \frac{\partial h}{\partial \phi} \\        
    \end{bmatrix}^T = 
    \begin{bmatrix}
        \cos \theta \\
        \sin \theta \\
        r_m \mu \sin \beta_\mu + r_m \mu v_H/v_A        
    \end{bmatrix}^T \\ \label{eq: delta h}
    \nabla h \dot x =&
    \begin{bmatrix}
        \cos \theta \\
        \sin \theta \\
        r_m \mu \sin \beta_\mu + r_m \mu v_H/v_A        
    \end{bmatrix}^T 
    \begin{bmatrix}
        v_A \cos \phi  \\
        v_A \sin \phi\\
        \omega
    \end{bmatrix}\\
    =& v_A (\cos \theta \cos \phi  + \sin \theta \sin \phi)\\
    &+\omega r_m \mu (\sin \beta_\mu + v_H/v_A) \\
    =& -v_A \sin \beta_\mu + (\mu \omega_m) r_m \mu (\sin \beta_\mu + v_H/v_A) \\
    =& -v_A \sin \beta_\mu + v_A (\sin \beta_\mu + v_H/v_A)\\
    =& v_H
\end{align}
Since we have $\frac{\partial h}{\partial t}= -v_H$ we get
$\dot{h}(\mathbf{x},\mathbf{u})= \nabla h \dot x+ \frac{\partial h}{\partial t} =0$ and thus \eqref{eq: h_lim}  holds.

Above we have used $\omega = \mu \omega_m$, 
$\mu^2 =1$, $\omega_m r_m = v_A$ and the following identity:
$        v_A (\cos \theta \cos \phi_A  + \sin \theta \sin \phi_A) = v_A \cos (\theta - \phi_A) 
        = v_A \cos(\mu(\theta - \phi_A)) 
        = -v_A \sin (\mu(\theta - \phi_A) - 0.5 \pi) 
        = -v_A \sin \beta_\mu$.
    
\end{proof}

\begin{theorem} [Exact safe set] \label{theorem: move exact safe set}
    Consider the CBF $h_\mu$ in Theorem \ref{theorem: move}, and let $\mathcal{C} := \mathcal{C}_{-1} \cup \mathcal{C}_1$. The set $\mathcal{C}$ is then the maximal control invariant set with respect to the halfplane: starting inside $\mathcal{C}$ is necessary and sufficient for the existence of an admissible input sequence that avoids the plane for all future time.
\end{theorem}

\begin{proof}
    For sufficiency, note that if $h_\mu \geq 0$ for some $\mu$, then by geometry, there exists the least-intrusive minimum-radius avoidance maneuver in direction $\mu$ that reaches a heading 
    where the closing speed is zero,
    without breaching the plane. This is because of the clearance term \eqref{eq: b for moving h}, which is a lower bound on the space needed to avoid the halfplane. Hence, a feasible control law exists that keeps the state in $\mathcal{C}_\mu$.

    For necessity, note that if $h_\mu < 0$ for both $\mu$, then a minimum radius avoidance maneuver in either direction will breach the halfplane. Hence, there exists no admissible control $|\omega| < \omega_m$ that can keep the system safe. Thus, no controller can keep the system safe, implying that the current state lies outside the maximal invariant set. Therefore, $\mathcal{C}$ is exactly the maximal control invariant set.
\end{proof}

\begin{remark}
    We note that by combining the two CBFs $h_{-1}$ and $h_1$, it is possible to ensure safety on $\mathcal{C}$ as long as at least one CBF constraint is satisfied.
\end{remark}

\subsection{Special Case: Stationary halfplane}

In the special case when the halfplane is stationary, $\mathbf{v}_O = 0$, the CBF in Theorem \ref{theorem: move} can be simplified.

\begin{lemma}
    If $\mathbf{v}_O = 0$ in Theorem \ref{theorem: move}, then $b$ in \eqref{eq: b for moving h} simplifies to
    \begin{equation} \label{eq: b for stat h}
        b = r_m (1 - \cos \beta_\mu).
    \end{equation}
\end{lemma}

\vspace{2mm}
\begin{proof}
    By letting $\gamma = 0$ and $v_H = 0$ in \eqref{eq: b for moving h}, we get \eqref{eq: b for stat h}.
\end{proof}

\begin{figure}[h!]
    \centering
    \def\svgwidth{0.49\textwidth}\footnotesize
    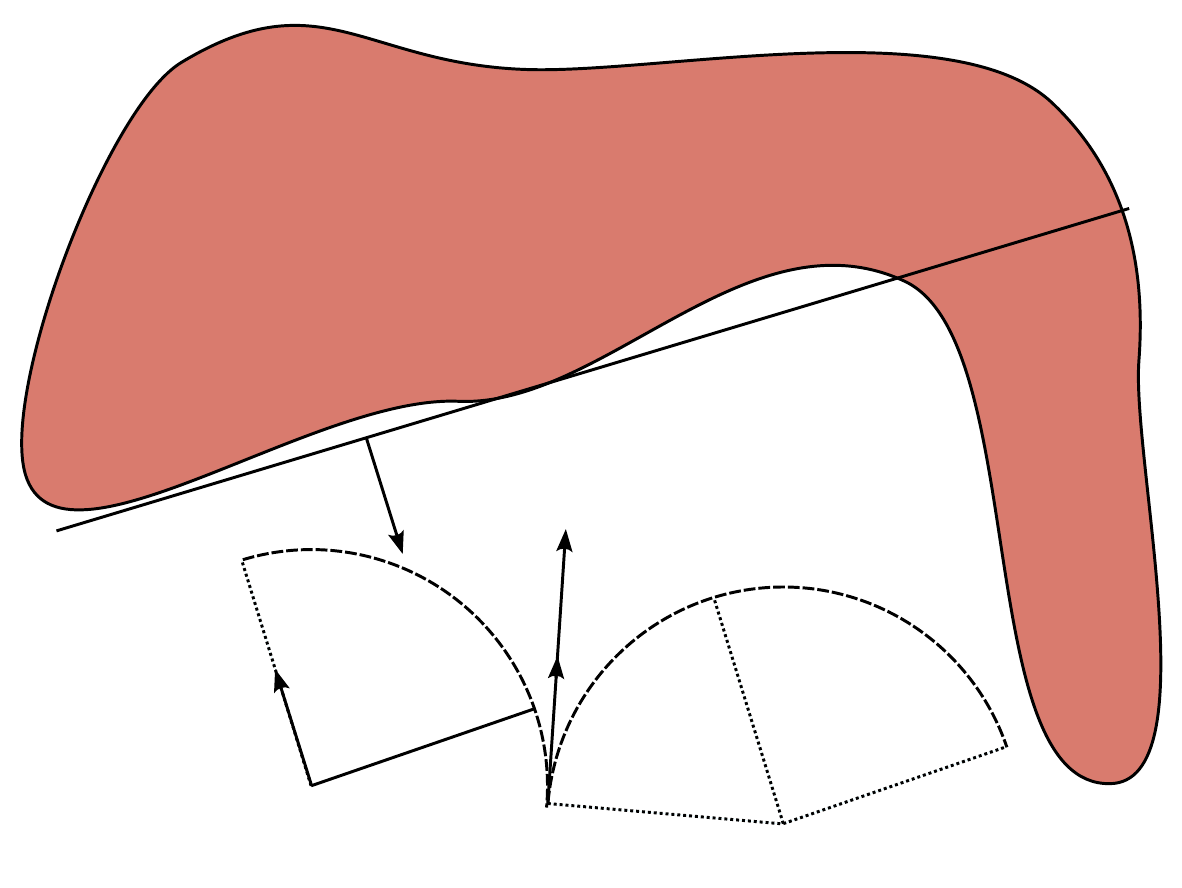
    \caption{The geometry of an agent, $A$, moving towards a corner of two stationary halfplanes, $O_1$ and $O_2$. Note that only the longest turn in each direction will ensure safety, corresponding to the requirement of Lemma \ref{lemma: corner}.}
    \label{fig: corner geometry}
\end{figure}

\subsection{Multiple halfplanes} \label{subsection: Multiple halfplanes}

With non-convex obstacle configurations, it can be useful to describe the unsafe set as the combination of multiple halfplane CBFs. To ensure recursive feasibility, we provide the following lemma: 

\begin{lemma} \label{lemma: corner}
    With two or more halfplane CBFs, it is sufficient for safety to find an $\omega$ such that all H-CBFs in one turning direction ($\mu = 1$ or $\mu = -1$) are satisfied.
\end{lemma}

\begin{proof}
    If there exists an $\omega$ that satisfies all H-CBFs in one turning direction, then this turn will be safe. See the illustration in Figure \ref{fig: corner geometry}.
\end{proof}

\begin{remark}
    While Lemma \ref{lemma: corner} ensures safety, it might seem conservative.
    For example, it will not allow the agent to enter corridors that are too narrow to afford a complete U-turn. However, this is reasonable when relying on only local information and having a vehicle without brakes.
\end{remark}

 \begin{figure}[h!]
        \centering
        \def\svgwidth{0.4\textwidth}\footnotesize
\begingroup%
  \makeatletter%
  \providecommand\color[2][]{%
    \errmessage{(Inkscape) Color is used for the text in Inkscape, but the package 'color.sty' is not loaded}%
    \renewcommand\color[2][]{}%
  }%
  \providecommand\transparent[1]{%
    \errmessage{(Inkscape) Transparency is used (non-zero) for the text in Inkscape, but the package 'transparent.sty' is not loaded}%
    \renewcommand\transparent[1]{}%
  }%
  \providecommand\rotatebox[2]{#2}%
  \newcommand*\fsize{\dimexpr\f@size pt\relax}%
  \newcommand*\lineheight[1]{\fontsize{\fsize}{#1\fsize}\selectfont}%
  \ifx\svgwidth\undefined%
    \setlength{\unitlength}{566.92913386bp}%
    \ifx\svgscale\undefined%
      \relax%
    \else%
      \setlength{\unitlength}{\unitlength * \real{\svgscale}}%
    \fi%
  \else%
    \setlength{\unitlength}{\svgwidth}%
  \fi%
  \global\let\svgwidth\undefined%
  \global\let\svgscale\undefined%
  \makeatother%
  \begin{picture}(1,0.65)%
    \lineheight{1}%
    \setlength\tabcolsep{0pt}%
    \put(0,0){\includegraphics[width=\unitlength,page=1]{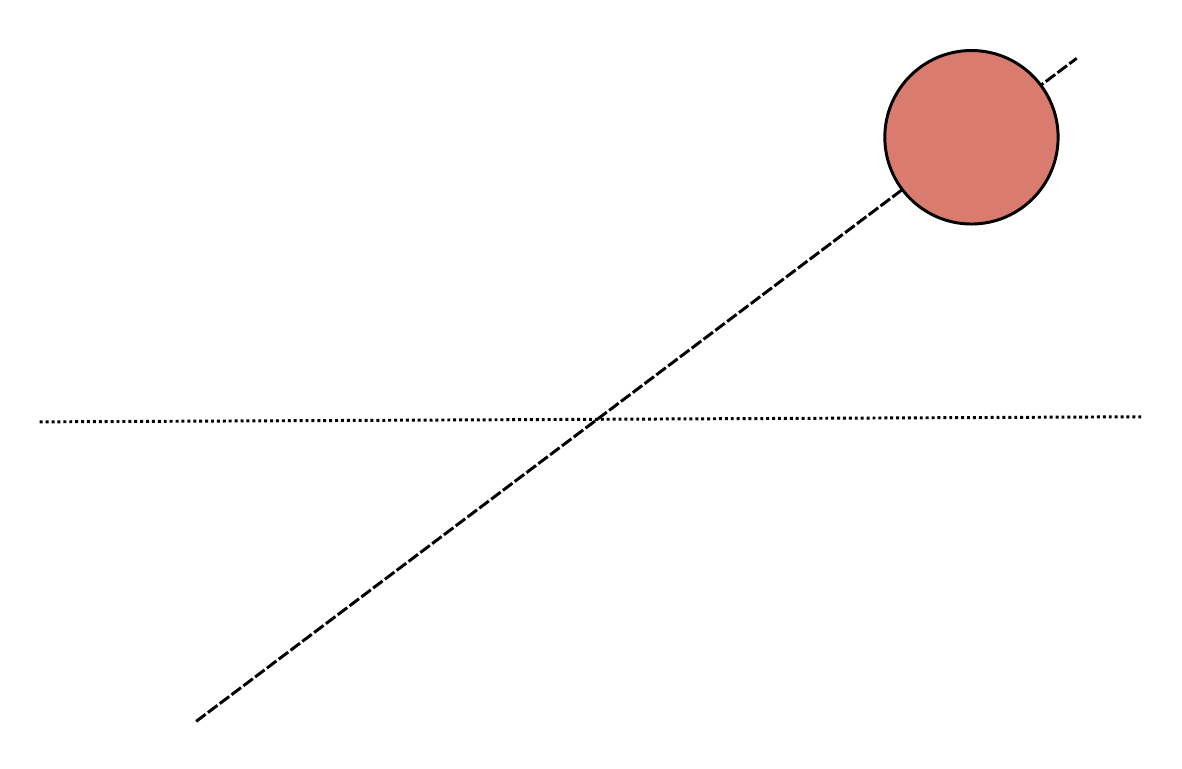}}%
    \put(0.14816975,0.25004145){\color[rgb]{0,0,0}\makebox(0,0)[lt]{\lineheight{1.25}\smash{\begin{tabular}[t]{l}$\mathbf{p}_A$\end{tabular}}}}%
    \put(0.86513361,0.45176275){\color[rgb]{0,0,0}\makebox(0,0)[lt]{\lineheight{1.25}\smash{\begin{tabular}[t]{l}$\mathbf{p}_O$\end{tabular}}}}%
    \put(0.71758063,0.41419891){\color[rgb]{0,0,0}\makebox(0,0)[lt]{\lineheight{1.25}\smash{\begin{tabular}[t]{l}$\mathbf{v}_O$\end{tabular}}}}%
    \put(0.25244029,0.32722452){\color[rgb]{0,0,0}\makebox(0,0)[lt]{\lineheight{1.25}\smash{\begin{tabular}[t]{l}$\mathbf{v}_A$\end{tabular}}}}%
    \put(0,0){\includegraphics[width=\unitlength,page=2]{Astroids.pdf}}%
    \put(0.59401337,0.31932395){\color[rgb]{0,0,0}\makebox(0,0)[lt]{\lineheight{1.25}\smash{\begin{tabular}[t]{l}$\psi$\end{tabular}}}}%
    \put(0,0){\includegraphics[width=\unitlength,page=3]{Astroids.pdf}}%
  \end{picture}%
\endgroup%

        \caption{
        The geometry of Example \ref{ex: astro}. The agent starts at $[-30, 0]^T$, with a goal at $[30, 0]^T$, moving at $v_A = 0.8$ m/s. The obstacle is parametrised by the incidence angle $\psi$ and the velocity $v_O$. The initial position of the obstacle is set so that it intersects the agent's nominal trajectory at $\mathbf{x} = [0, 0]^T$. The simulation is run for up to $300$ s, or until the agent is within $1$ m of the goal. The nominal time to goal is $60 / 0.8 = 75$ s.
        }
        \label{fig: astroids}
    \end{figure}

\section{Simulations}
\label{section: Experiments}

In this section, the theoretical results from Section~\ref{section: Method} will be illustrated in two simulation examples, while the hardware experiment is presented in the next section. 
First, we evaluate the proposed approach compared to the state-of-the-art in a scenario where the agent encounters a moving obstacle.  
We collect statistics by running a set of simulations with different obstacle velocities and incidence angles for each method.
Second, the agent encounters a non-convex obstacle configuration, in the form of a harbour, where it uses multiple halfplanes to safely navigate through a set of waypoints.



    \begin{figure}[!h]
        \centering
        \def\svgwidth{0.49\textwidth}\footnotesize
        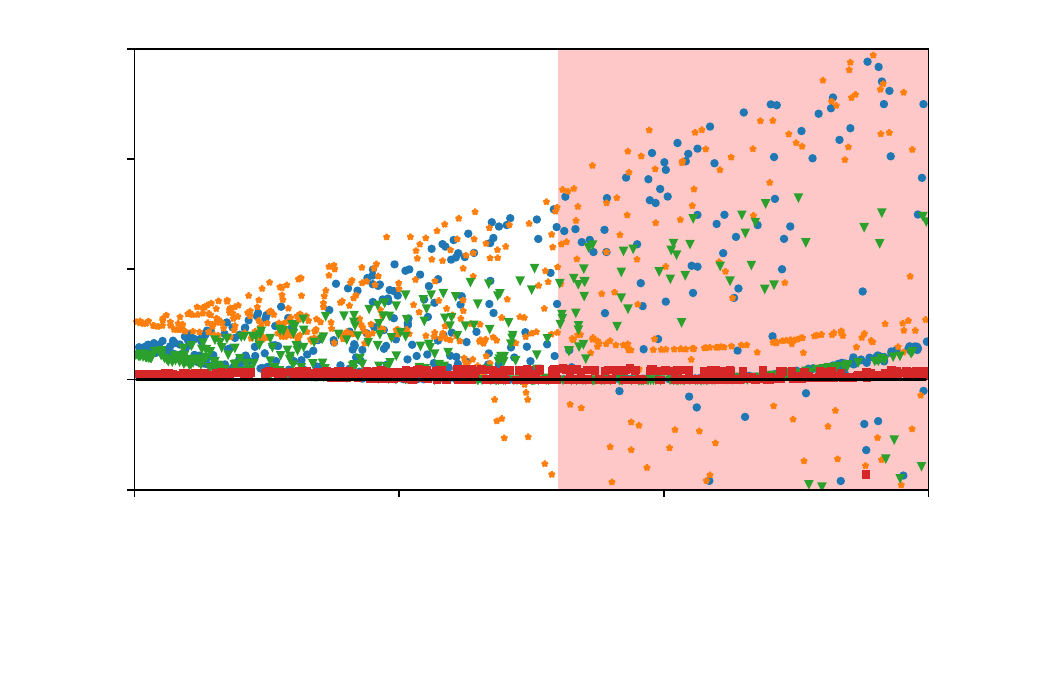
        \caption{
        The closest distance between the agent and the obstacle, as a function of the speed of the obstacle. 
        Notice how the LRH-CBF allows the agent to go close to the obstacle consistently. 
        All methods ensure safety for all $v_O < v_A = 0.8$ m/s, with the exception of the SHO-CBF, which has a few collisions at obstacle speeds of $v_O \in (0.65, 0.8)$.
        }
        \label{fig: astro obs speed}
    \end{figure}

    \begin{figure}[!h]
        \centering
        \def\svgwidth{0.49\textwidth}\footnotesize
        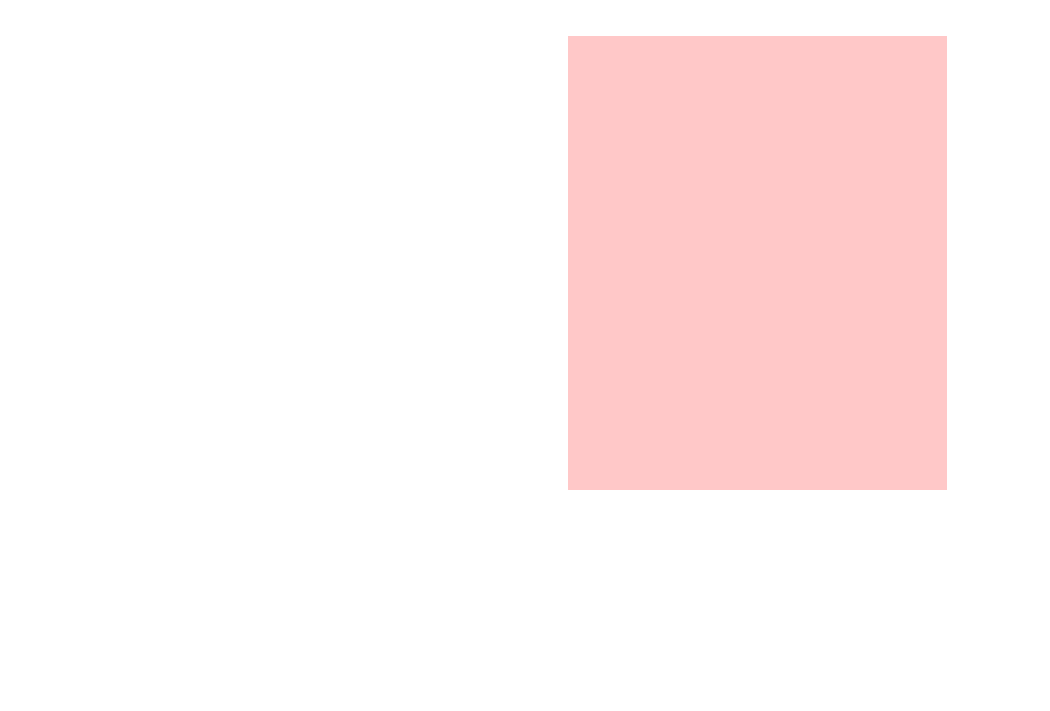
        \caption{
        The time delay to the goal, compared to the direct trajectory, as a function of the speed of the obstacle. 
        The y-axis is on a log scale to better show differences over a large range of values. 
        The lines are the moving average over a window of 51 points. 
        Notice how the LRH-CBF has significantly smaller delays than all other methods, with MDH-CBF delays often 50 \% larger and (S)HO-CBF even larger still.
        }
        \label{fig: astro goal time}
    \end{figure}

\begin{table}[h!]
\centering
\caption{Average Obstacle Distances and Arrival Delays. Note that without the moving obstacle, the straight-line distance is 60m with a corresponding travel time of 75 s, which is used to compute the delays.}
\label{tab:results}
\begin{tabular}{|c|c|c|}
\hline
\textbf{Method:} & \textbf{Closest Distance (avg) [m]} & \textbf{Delay (avg) [s]} \\ \hline
HO-CBF & $0.817 \pm 1.33$ & $34.5 \pm 52.0$ \\ \hline
SHO-CBF & $0.650 \pm 2.55$ & $33.3 \pm 51.2$ \\ \hline
MDH-CBF & $2.03 \pm 1.53$ & $10.3 \pm 5.14$ \\ \hline
LRH-CBF (ours) & $\mathbf{0.105} \pm 0.359$ & $\mathbf{6.79} \pm 3.29$ \\ \hline
\hline
\end{tabular}
\end{table}

\begin{example} \label{ex: astro}

    The first experiment verifies that the proposed halfplane CBF for Dubins car dynamics works as expected and compares it with the state-of-the-art. The main competitor is Synergetic HO-CBF (SHO-CBF) from \cite{haraldsen_hybrid_2025}, which was recently proposed for obstacle avoidance for boats. To avoid picking tuning parameters for their solution, we will copy their obstacle and vehicle parameters for the experimental setup, combined with their SHO-CBF parameters. We will also include a conventional HO-CBF, tuned with the same parameters as the SHO-CBF.
    The LRH-CBF uses $\alpha(x) = 0.1 x$ and 12 discrete halfplanes: one of these 12 is the previous one (needed for recursive feasibility, see \cite{trende_least_2026}), and the other 11 are uniformly distributed on $\theta \in [\theta_{OA}-0.4 \pi, \theta_{OA} + 0.4 \pi]$, where $\hat{\mathbf{n}}(\theta_{OA})$ is the unit vector pointing from the obstacle towards the agent. We also compare with an MDH-CBF version of our proposed approach, with $\theta = \theta_{OA}$.
    
    The experimental setup is illustrated in Figure \ref{fig: astroids}:   
    The agent starts at $x = -30$ m and drives toward the goal at $x = 30$ m using a PD controller on the heading error to the goal. 
    Using parameters from \cite{haraldsen_hybrid_2025},
    the agent's speed is constant and set to $v_A = 0.8$ m/s, with $\omega_\text{max} = 0.5$ rad/s. The agent is modeled as a disc with radius 1 m, while the obstacle is a disc with radius 9 m, moving at a constant velocity $v_O$ at an incidence angle $\psi$, configured such that it intersects the nominal path of the agent at $\mathbf{x} = [0, 0]^T$ m at the agent's nominal arrival time of $t = 37.5$ s. The simulation is run for up to 300 s or until the agent reaches within 1 m of the goal.

    The simulation was run 500 times for each CBF type, drawing random values uniformly for $v_O \in [0, 1.5]$ and $\psi \in (-\pi, \pi]$, resampling if the initial condition yields $h(\mathbf{x}) < 0$. The results are presented in Figure \ref{fig: astro obs speed}, where the closest distance to the obstacle during each simulation is plotted against the obstacle speed. 
    All CBFs, with the exception of the SHO-CBF, which has a few collisions for higher obstacle speeds, ensure safety for all obstacle angles and obstacle velocities smaller than the vehicle, $v_O < v_A = 0.8$ m/s. For the MDH-CBF and LRH-CBF, this corresponds to the safety guarantees we expect from Theorems \ref{theorem: move} and \ref{theorem: move exact safe set}. We also include significantly higher obstacle speeds to see how the filters perform outside their nominal operating envelope.
    
    Safety is important, but getting to the goal is also an objective. In Figure \ref{fig: astro obs speed}, we see that the LRH-CBF is allowed to get closer to the obstacle than the other three methods. When we measure the time delay in getting to the goal, presented in Figure \ref{fig: astro goal time}, we see that this translates to a much smaller delay for the LRH-CBF. 
    
    The average results are found in Table \ref{tab:results}. As can be seen, LRH-CBF produces an average delay of 6.79 seconds, compared to a direct trajectory, while the second-best MDH-CBF has a delay of 10.3 seconds and the other two approaches give delays above 30 seconds. Thus, the LRH-CBF reduces delays by more than 33 \% in this scenario.
    

\end{example}

\begin{example} \label{ex: mace}
    In this example, an agent follows a series of waypoints in a harbour environment, as shown in Figure~\ref{fig: mace}. This scenario illustrates the agent's ability to maneuver safely in non-convex environments by using multiple halfplanes. The agent is modeled as a Dubins car with $v_A = 5$ m/s, $\omega_\text{max} = 0.5$ rad/s, and the LRH-CBF uses 12 halfplanes and $\alpha(x) = x$. The obstacles are divided into convex parts, each of which is assigned a halfplane. As seen in Figure \ref{fig: mace}, a safe passage to the goal is ensured.
    
    \begin{figure}[!t]
        \centering
        \def\svgwidth{0.49\textwidth}\footnotesize
        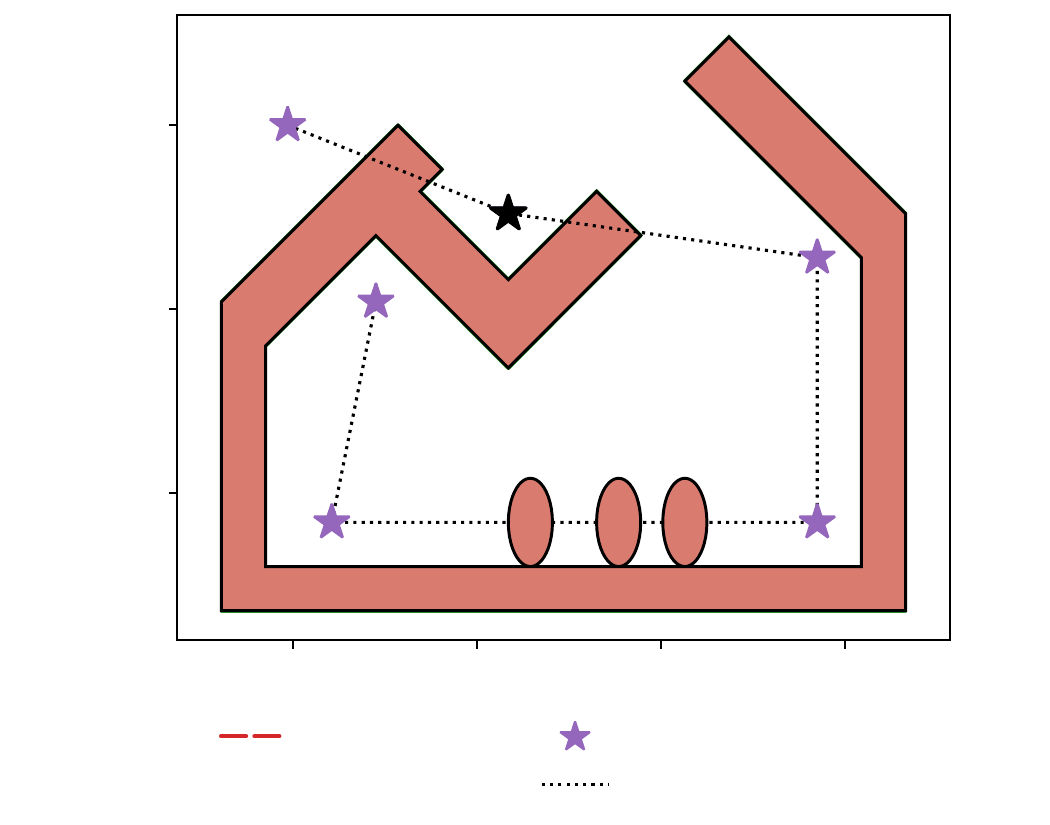
        \caption{The agent drives in a harbour, following a series of waypoints. By using multiple halfplanes to describe the environment, the agent's safety can be ensured while it fulfills its mission.}
        \label{fig: mace}
    \end{figure}
\end{example}

\section{Evolo Experiments}
\label{section: Evolo}

To validate the proposed approach on a complex Autonomous Surface Vehicle (ASV), we conducted sea trials\footnote{A video from the sea trials can be found at: https://drive.google.com/file/d/1H6EIisT3DlfIOAjDK0KwBsD-nufFnHZB/view?usp=sharing} with our autonomous hydrofoiling boat, Evolo.
For mechanical details about Evolo, see \cite{j_kuttenkeuler_evolo_2026}. To model Evolo as a Dubins car, we use the parameters $v_A = 5$ m/s and $\omega_m = 0.5$ rad/s, which correspond to a turning radius of $10$ m. The spatial extent of Evolo is modeled as a circle with radius $1$~m. The LRH-CBF uses 12 halfplanes and $\alpha(x) = x$.

\begin{figure}[!t]
    \centering
    \def\svgwidth{0.49\textwidth}\footnotesize
    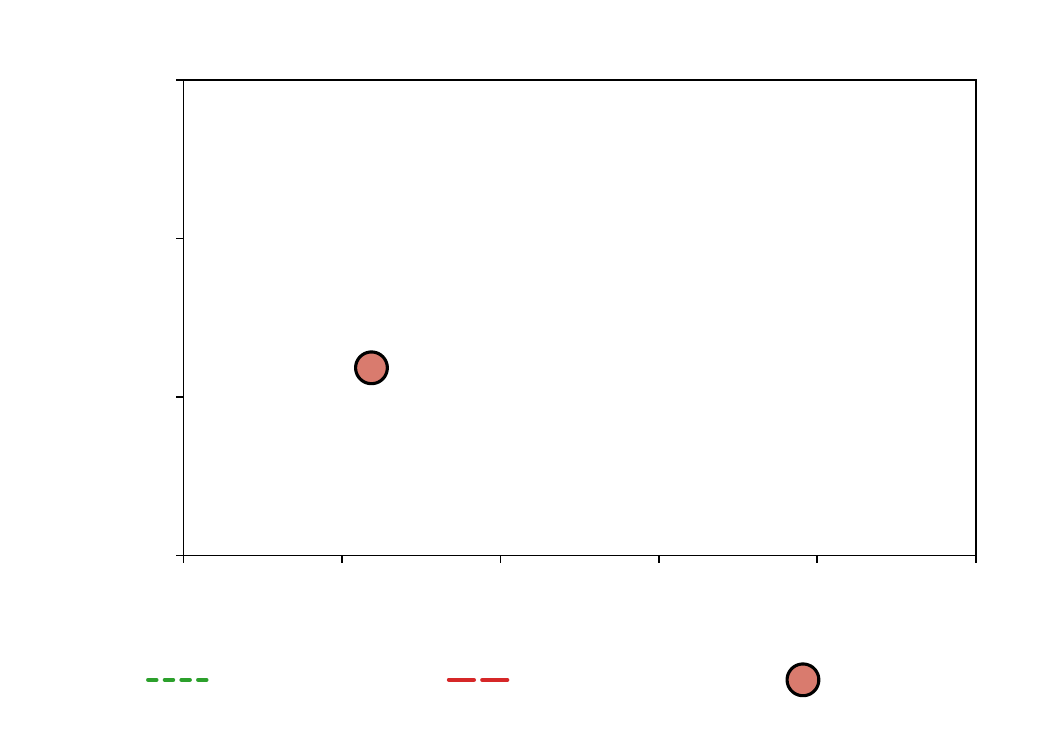
    \caption{Evolo's paths for the MDH-CBF and the LRH-CBF for a) a static obstacle, b) a moving obstacle with $v = 1$ m/s. The closest instance between Evolo and the moving obstacle occurs at $t = 9.8$ s, and a snapshot of both Evolo and the obstacle at that time is shown.}
    \label{fig: evolo results}
\end{figure}

To facilitate the comparison between the simulation and the real vehicle, we chose a similar setup to Example \ref{ex: astro}. In the first run, a static obstacle with radius $9$ m was placed on the nominal path as Evolo is driving on a straight line. In the second run, an obstacle moving at 1 m/s at $\psi = \pi / 2$ intersects the nominal path. The results for both runs are presented in Figure \ref{fig: evolo results}. 

Two main features of the hardware experiments can be considered. First, we note that the CBF applied to the real vehicle is more conservative with regard to the obstacles compared to the simulations. This is likely due to model error, where the low-level controllers have a slightly non-linear response to the requested $\omega$. Second, we note that the qualitative behaviour differences between the MDH-CBF and the LRH-CBF are the same. With the LRH-CBF, a smaller deviation from the desired path is achieved, and Evolo is able to move closer to the obstacle while staying safe, thus reaching the goal faster.

\begin{remark}
Moving closer to obstacles might not sound like a good choice when deploying an autonomous system on the water, especially not when there are known modelling errors. However, with a system that is known to be less conservative, we can make an estimate of how much margin we need due to the modelling errors, and add that through the parameter $\delta$ in \eqref{eq: b for moving h}.
Whereas for a system that is already conservative, but in a way that is hard to predict, it is harder to know how much extra margin is needed.
Either way, going deeper into the question of modelling errors is beyond the scope of this paper.
\end{remark}

\section{Conclusion}
\label{section: Conclusion}

In this paper, we proposed a CBF for Dubins car models and applied it to a hydrofoiling boat.
In numerical simulations, we showed that it was less conservative and gave shorter travel times than earlier approaches.

In future work, we will investigate how to extend the approach to account for proper adherence to the traffic rules of the sea, known as COLREGs.



\addtolength{\textheight}{0cm}   


\bibliographystyle{IEEEtran}
\bibliography{Dubin_CBF}

\end{document}